\documentclass[a4paper]{article}

\usepackage{INTERSPEECH2018}
\usepackage{xcolor}
\usepackage{multirow}
\usepackage{footnote}
\makesavenoteenv{tabular}
\makesavenoteenv{table}
\usepackage{enumitem}
\usepackage{flushend}
\usepackage{hhline}

\title{End-to-End Speech Recognition From the Raw Waveform}
\name{Neil Zeghidour$^{1,2}$, Nicolas Usunier$^{1}$, Gabriel Synnaeve$^{1}$, Ronan Collobert$^{1}$, Emmanuel Dupoux$^2$}
\address{
$^1$ Facebook A.I. Research, Paris, France; New York \& Menlo Park, USA\\
$^2$ CoML, ENS/CNRS/EHESS/INRIA/PSL Research University, Paris, France}
\email{\{neilz, usunier, gab, locronan\}@fb.com, emmanuel.dupoux@gmail.com}

\begin{document}

\maketitle
\begin{abstract}
State-of-the-art speech recognition systems rely on fixed, hand-crafted features such as mel-filterbanks to preprocess the waveform before the training pipeline. 
In this paper, we study end-to-end systems trained directly from the raw waveform, building on two alternatives for trainable replacements of mel-filterbanks that use a convolutional architecture. The first one is inspired by gammatone filterbanks  (Hoshen et al., 2015; Sainath et al, 2015), and the second one by the scattering transform (Zeghidour et al., 2017).
We propose two modifications to these architectures and systematically compare them to mel-filterbanks, on the Wall Street Journal dataset. The first modification is the addition of an instance normalization layer, which greatly improves on the gammatone-based trainable filterbanks and speeds up the training of the scattering-based filterbanks. The second one relates to the low-pass filter used in these approaches. These modifications consistently improve performances for both approaches, and remove the need for a careful initialization in scattering-based trainable filterbanks. In particular, we show a consistent improvement in word error rate of the trainable filterbanks relatively to comparable mel-filterbanks. It is the first time end-to-end models trained from the raw signal significantly outperform mel-filterbanks on a large vocabulary task under clean recording conditions.
\end{abstract}

\section{Introduction}

State-of-the-art speech recognition systems rapidly shift from the
paradigm of composite subsystems trained or designed independently to
the paradigm of end-to-end training. While most of the work in this
direction has been devoted to learning the acoustic model directly from
sequences of phonemes or characters without intermediate alignment
step or phone-state/senome induction, the other end of the pipeline
model -- namely, learning directly from the waveform rather than from
speech features such as mel-filterbanks or MFCC -- has recently
received attention
\cite{palaz_baseline,hermann,hoshen2015speech,sainath2015learning,zhu2016learning,wavenet,ghahremani2016acoustic,neilfbanks},
but the performances on the master task of speech recognition still
seem to be lagging behind those of models trained on speech features \cite{msr2017, capio2017}.

Yet, promising results have already been obtained by learning the
front-end of speech recognition systems. We focus the discussion on
trainable components that can be plugged in as replacement
of mel-filterbanks without modification of the acoustic model. The approach inspired by
gammatone filterbanks of Hoshen et al. and Sainath et al. \cite{hoshen2015speech, sainath2015learning}
achieved similar or better results than comparable mel-filterbanks on
multichannel speech recognition and on far-field/noisy recording
conditions. More recently, Zeghidour et al. \cite{neilfbanks} proposed an alternative
learnable architecture based on a convolutional architecture that
computes a scattering transform and can be initialized as an approximation of mel-filterbanks, and
obtained promising results on end-to-end phone recognition on TIMIT. However, these
approaches have not been proven to improve on speech features on
large-scale, end-to-end speech recognition in clean recording
conditions on English -- admittedly one of the tasks for which
mel-filterbanks have been the most extensively tuned.

We present a systematic comparison of the two previous
architectures of learnable filterbanks, which we will (coarsely) refer to
as gammatone-based and scattering-based, and evaluate them against
mel-filterbanks within an end-to-end training pipeline on letter error
rate and word error rate on the Wall Street
Journal dataset. Our main contributions and results are the following:
\begin{enumerate}[leftmargin=10pt]
\item A mean-variance normalization layer on
top of the log nonlinearity of learnable filterbanks appears to be critical
for the efficient learning of the gammatone-based architecture, and
makes the training of the scattering-based architecture faster;
\item The low-pass filter previously used in the
scattering-based learnable filterbanks stabilizes the training
of gammatone filterbanks, compared to the max-pooling that was originally proposed
\cite{hoshen2015speech,sainath2015learning};
\item
For scattering-based trainable filterbanks, keeping the
low-pass filter fixed during training allows to efficiently learn the
filters from a random initialization, whereas the results
of \cite{neilfbanks} with random initialization of both the filters
and the low-pass filter showed poor performances compared to a suitable
initialization;
\item Both trainable
filterbanks improve against the mel-filterbanks baseline on word error rate on the Wall Street Journal dataset, in similar conditions (same
number of filters, same end-to-end training convolutional
architecture). This is the first time learnable filterbanks improve against a strong mel-filterbanks
baseline on a large vocabulary, speech recognition task under clean recording conditions.
\end{enumerate}
The next section describes the learnable filterbanks architectures. Then, we present the end-to-end convolutional architecture used to perform the comparisons, and analyze the results of our comparative studies.

\begin{figure*}
  \begin{center}
    \includegraphics[width=1.7\columnwidth]{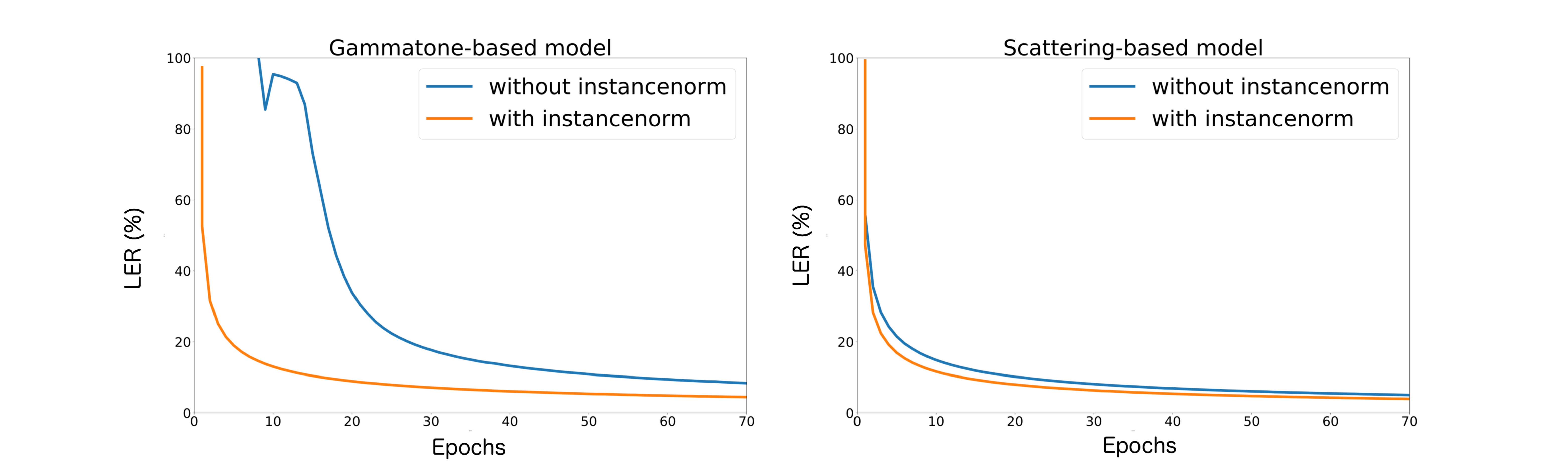}\hfill
    \caption{\label{fig:instancenorm} Training Letter Error Rate (LER) for a gammatone-based architecture (left) and a scattering-based architecture (right), with and without instance normalization.}
  \end{center}
\end{figure*}

\section{Learning filterbanks from raw speech}

The two approaches that we consider for learning filterbanks from the raw waveform can be used as direct replacement for mel-filterbanks in any end-to-end learning pipeline: they are convolutional architectures that take the raw waveform as input and output $40$ channels every $10$ms. As such, they can directly be compared with standard mel-filterbanks, simply by changing the features stage of a neural-network-based acoustic model. The filters are then nothing more than an additional layer to the neural network and are learnt by backpropagation with the rest of the acoustic model.

The first architecture we consider is inspired by \cite{hoshen2015speech,sainath2015learning}, the second one is taken from \cite{neilfbanks}. They are described in Table~\ref{tab:archis}. 

In both architectures, a convolutional layer with window length $25$ms (to match the standard frame size used in mel-filterbanks) is applied with a stride of $1$ sample, and is followed by a nonlinearity to give $40$ output channels for each sample. Then, a pooling operator of width $25$ms with a stride of $10$ms performs low-pass filtering and decimation. Finally, a log non-linearity reproduces the dynamic range compression  of log mel-filterbanks. The parameters to be learnt are the convolution filters, and possibly the weights of the low-pass filters.

The two architectures differ by the choices of each layer of computation. Hoshen et al. and Sainath et al. use $40$ real-valued filters with ReLU non-linearity, and rely on gammatones as filter values to approximate mel-filterbanks \cite{hoshen2015speech,sainath2015learning}. In their work, they use a max-pooling operator for low-pass filtering. In contrast, Zeghidour et al. \cite{neilfbanks} use $40$ complex-valued filters with a square modulus operator as non-linearity. Low-pass filtering is then performed by multiplying each output channel by a squared Hanning window so that, when using suitable Gabor wavelets as convolution filters, the architecture closely approximates mel-filterbanks computed on the power spectrum \cite{dss}.

The number of filters ($40$), the convolution and pooling width of $25$ms, as well as the decimation of $10$ms are not necessarily the optimal parameters of either trainable architecture, but these are the standard settings of mel-filterbanks (and likely the best settings for these features on standard speech recognition datasets). We keep these values fixed for the trainable architectures, so that the comparison to mel-filterbanks is carried out in the setting most favorable for the non-learnable baseline.

In the next subsections, we describe the improvements we propose for these architectures: the low-pass filter and the addition of instance normalization.

\subsection{Low-pass filtering}
\label{sec:lowpass}
The original papers describing the gammatone-based trainable filterbanks used max-pooling as low-pass filter, whereas the scattering-based approach uses a squared Hanning window per channel. To make sure the low-pass filter is not responsible for notable differences between the two approaches we experiment with the squared Hanning window on both architectures. For both architectures, we also propose to keep this low-pass filter fixed while learning the convolution filter weights, a setting that was not explored by Zeghidour et al. \cite{neilfbanks}, who learnt the low-pass filter weights when randomly initializing the convolutions.

\subsection{Instance normalization}
\label{sec:instancenorm}
More importantly, we noticed that a per-channel per-sentence mean-variance normalization after log-compression is important for the baseline mel-filterbanks. Consequently, we propose to add a mean-variance normalization layer on both trainable architectures, performed for each of the $40$ channels independently on each sentence. Coincidently, this corresponds to an instance normalization layer \cite{instancenorm}, which has been shown to stabilize training in other deep learning contexts.

\begingroup
\newcommand\Tstrut{\rule{0pt}{2.6ex}}         
\newcommand\Bstrut{\rule[-0.9ex]{0pt}{0pt}}   
\setlength{\tabcolsep}{6pt}
\begin{table}[t]
    \centering
    \begin{tabular}{lll}
    \toprule
    & {\sc scattering} & {\sc gammatones} \Tstrut\Bstrut \\
    \midrule
    Conv\footnote{The convolution for the scattering-based architecture uses $80$-real valued output channels and squared L2-pooling on the feature dimension to emulate a complex-valued convolution with $40$ filters followed by a squared modulus operator. Thus, after the nonlinearity, both architectures have $40$ filters.}  & \multirow{2}{*}{1-80-400-1} & \multirow{2}{*}{1-40-400-1} \\ 
    {\footnotesize \emph{(\#in-\#out-width-stride)}} & & \Bstrut\\ \hline
    non-linearity & sq. L2-Pooling & ReLU\Tstrut\Bstrut\\\hline
    low-pass filter & \multirow{2}{*}{sq. Hanning} & max-pooling \Tstrut\\
    {\footnotesize \emph{(wdth=$400$, strd=$160$)}}& & or sq. Hanning \Bstrut\\ \hline
    log-compression\footnote{\cite{neilfbanks} use $1$ to prevent $\log(0)$ and \cite{hoshen2015speech,sainath2015learning} use $0.01$. We kept the values initially used by the authors of the respective papers and did not try alternatives. We believe it has little impact on the final performance.} & $\log(1 \!+\!\mathop{{\rm abs}}(.))$ & $\log(0.01\! + \!\mathop{\rm abs}(.))$\Tstrut\Bstrut\\ \hline
    normalization & \multicolumn{2}{c}{mean-var. per-channel per-sentence} \Tstrut\\ 
    \bottomrule
    \end{tabular}
    \caption{Architectures of the two trainable filterbanks. Values of width and strides are given to match the standard settings of mel-filterbanks for waveform sampled at $16$kHz.}
    \label{tab:archis}
\end{table}
\endgroup

\section{Experimental setup}

The experiments compare different versions of the trainable architectures against log mel-filterbanks on a single deep convolutional network architecture for the acoustic model. The experiments are carried out on the open vocabulary task of the Wall  Street  Journal dataset \cite{wsj}, using the subset si284 for training, nov93-dev for validation, and nov92-eval for testing. Training is performed end-to-end on letters. We evaluate in both letter and word error rates. All our experiments use the open source code of wav2letter \cite{wav2letter}. In the next subsections, we describe the model, the different variants we tested and the hyperparameters.

\subsection{Acoustic model}

Taking either log mel-filterbanks or trainable filterbanks, the acoustic model is a convolutional network with gated linear units (GLU) \cite{glu} trained to predict sequences of letters, following \cite{baseline}. The model is a smaller version of the convolutional network used in \cite{baseline} since they train on the larger LibriSpeech dataset. Using the syntax C-input channels-output channels-width,
the architecture we use has the structure C-40-200-13/C-100-200-3/C-100-200-4/C-100-250-5/\\C-125-250-6/C-125-300-7/C-150-350-8/C-175-400-9/\\C-200-450-10/C-225-500-11/C-250-500-12/C-250-500-13/\\C-250-600-14/C-300-600-15/C-300-750-21/C-375-1000-1. All convolutions have stride $1$. The number of input channels of the $n+1$th convolution is half the size of the output of the $n$-th convolution because of the GLU. There are GLU layers with a dropout \cite{dropout} of $0.25$ after each convolution layer. There is an additional linear layer to predict the final letter probabilities. When predicting letters, the training and decoding are performed as in \cite{baseline}. When predicting words, we use a 4-gram language model trained on the standard LM data of WSJ \cite{wsj} and perform beam search decoding, as in \cite{baseline}.

\subsection{Variants}

We compare the two architectures of trainable filterbanks along different axes: how to initialize the convolutions of the trainable filterbanks, the low-pass filter, and instance normalization.
\subsubsection{Gammatone-based architecture}
\begin{description}[leftmargin=0pt]
\item[Initialization of the convolution weights] random (rand), or with gammatone filters (gamm) that match the impulse response of a reference open source implementation of gammatones \cite{gammatones};
\item[Low-pass filter] max-pooling as in \cite{hoshen2015speech}, or the squared Hanning window (Han-fixed).
\end{description}
\subsubsection{Scattering-based architecture}
\begin{description}[leftmargin=0pt]
\item[Initialization of the convolution weights] random (rand), or Gabor filters (scatt) as described in Section 2.2 of \cite{neilfbanks};
\item[Low-pass filter] the squared Hanning window (Han-fixed), or a low-pass filter of same width and stride initialized with the weights of the squared Hanning window but the weights are then learnt by backpropagation (Han-learnt).
\end{description}

\subsection{Hyperparameters and training}
For models trained on the raw waveform, the signal was first normalized with mean/variance normalization by sequence.  The network is trained with stochastic gradient descent and weight normalization \cite{salimans2016weight} for all convolutional layers except the front-ends. First, $80$ epochs are performed with a learning rate of $1.4$, then training is resumed for $80$ additional epochs with a learning rate of $0.1$. These hyperparameters were chosen from preliminary experiments as they seemed to work well for all architectures. Additional hyperparameters are the momentum and the learning rate for the training criterion, respectively chosen in $\{0, 0.9\}$ and $\{0.001, 0.0001\}$ \cite{wav2letter,baseline}. 

For Letter Error Rate (LER) evaluations, the hyperparameters are selected using the LER on the validation set, validating every epoch. For Word Error Rate (WER) evaluations, the hyperparameters are chosen on the validation set using the WER, validating every $10$ epochs. The model selected on LER is also included for validation. The additional hyperparameters are the weight of the language model and the weight of word insertion penalty  (see \cite{baseline} for details). We set them between $5$ and $8$ by steps of $0.5$, and between $-2$ and $0.5$ by steps of $0.1$, respectively. For hyperparameter selection, the beam size of the decoder is set to $2,500$; the final performances are computed with the selected hyperparameters but using a beam size of $25,000$.

\section{Experiments}

\begingroup
\setlength{\tabcolsep}{3pt}
\begin{table}[t]
    \centering
\begin{tabular}{p{1.1cm}llcccc}\toprule\\ [-1.5ex]
\multicolumn{3}{c}{\multirow{2}{*}{\sc model}}&\multicolumn{2}{c}{\sc nov93-dev}&\multicolumn{2}{c}{\sc nov92-eval}\\
&&&{\sc LER}&{\sc WER}&{\sc LER}&{\sc WER}\\
\\ [-1.5ex]\hline\\ [-1.5ex]
\multicolumn{7}{c}{\small SOTA -- speech features}\\
\\ [-1.5ex]\hline\\ [-1.5ex]
\multicolumn{3}{p{3.8cm}}{Deep Speech 2 \cite{amodei2016deep}}&$-$&$-$&$-$&$3.6$\\
\multicolumn{3}{p{3.8cm}}{ -- {\small \emph{(+ additional data)}}}& & & &\\
\multicolumn{3}{p{3.8cm}}{RNN-WER - tri. LM \cite{graves2014towards}}&$-$&$-$&$-$&$8.2$\\
\multicolumn{3}{p{3.8cm}}{RNN - WSFT decoding \cite{miao2015eesen}}&$-$&$-$&$-$&$7.3$\\
\multicolumn{3}{p{3.8cm}}{Seq2Seq + tri. LM \cite{chorowski2016seqseq}}&$-$&$9.7$&$-$&$6.7$\\
\multicolumn{3}{p{3.8cm}}{Multi-task CTC/att \cite{kim2017joint}}&$11.3$&$-$&$7.3$&$-$\\
\multicolumn{3}{p{3cm}}{Att + RL \cite{tjandra2017sequence}}&$-$&$-$&$6.1$&$ $\\
\\ [-1.5ex]\hline\\ [-1.5ex]
\multicolumn{7}{c}{\small SOTA -- waveform}\\
\\ [-1.5ex]\hline\\ [-1.5ex]
\multicolumn{3}{p{3.8cm}}{Att Wav2Text (+transfer) \cite{tjandra2017attention}}&$-$&$-$&$6.5$&$-$\\
\multicolumn{3}{p{3.8cm}}{gamm {\footnotesize (learnt)/gamm/max-pool}}&$8.9$&$12.9$&$6.4$&$8.8$\\
\multicolumn{3}{p{3.8cm}}{ -- {\small \emph{(without inst. norm.)}}}& & & &\\
\\ [-1.5ex]\hhline{=======}\\ [-1.5ex]
{\centering {\sc front}}&{\centering {\sc filter}}&{\centering {\sc low-}}&\multicolumn{2}{c}{\sc nov93-dev}&\multicolumn{2}{c}{\sc nov92-eval}\\
{\centering {\sc end}}&{\centering {\sc init}}&{\centering {\sc pass}}&{\sc \small LER}&{\sc \small WER}&{\sc \small LER}&{\sc \small WER}\\
\\ [-1.5ex]\hline\\ [-1.5ex]{mel-fbanks} & & &  $6.9$ & $9.5$ & $4.9$ & $6.6$ \\
\\ [-1.5ex]\hline\\ [-1.5ex]\multirow{4}{1.1cm}{gamm\\(learnt)} & \multirow{2}{*}{gamm} & {Han-fixed} & $6.9$ &$9.1$&$4.9$&$5.9$\\
 &  & {max-pool} &$7.2$&$9.3$&$4.9$&$6.0$\\
 & \multirow{2}{*}{rand} & {Han-fixed} & $7$ &$8.9$ &$4.9$&$5.9$\\
 &  & {max-pool} & $7.2$ &$9.2$&$5.1$&$6.3$\\
\\ [-1.5ex]\hline\\ [-1.5ex]\multirow{4}{1.1cm}{scatt\\(learnt)} & \multirow{2}{*}{scatt} & {Han-fixed} &$6.7$&$8.3$&$4.6$&$6.1$\\
 &  & {Han-learnt} &$6.7$&$8.9$&$4.5$&$6.3$\\
 & \multirow{2}{*}{rand} & {Han-fixed} &$6.8$&$8.5$&$4.7$&$5.7$\\
 &  & {Han-learnt} &$6.9$&$8.9$&$4.9$&$5.8$\\
\bottomrule\end{tabular}
   \caption{Results on the open vocabulary task of the WSJ dataset.  \emph{(i)} SOTA -- speech features: for state-of-the-art and representative baselines using speech features (mel-filterbanks, spectrograms or MFCC), \emph{(ii)} SOTA-waveform: state-of-the-art from the raw waveform, including our own implementation of vanilla gammatone filterbanks without instance normalization, and \emph{(iii)} our baseline and the different variants of the trainable filterbanks (with instance normalization) studied in this paper.}
    \label{tab:results}
\end{table}
\endgroup

\begingroup
\setlength{\tabcolsep}{5pt}
\begin{table}[t]
    \centering
\begin{tabular}{p{1.1cm}lcccc}\toprule\\ [-1.5ex]
\multirow{2}{*}{\sc model}&\multirow{2}{*}{\sc pre-emp}&\multicolumn{2}{c}{\sc nov93-dev}&\multicolumn{2}{c}{\sc nov92-eval}\\
&&{\sc LER}&{\sc WER}&{\sc LER}&{\sc WER}\\
\\ [-1.5ex]\hline\\ [-1.5ex]\multirow{2}{0.55cm}{gamm\\(learnt)} & {no pre-emp} & $6.9$ &$9.1$&$4.9$&$5.9$\\
 & {pre-emp} & $6.8$ & $9$ & $4.7$ & $5.7$ \\
\\ [-1.5ex]\hline\\ [-1.5ex]\multirow{2}{0.55cm}{scatt\\(learnt)} & {no pre-emp} &$6.7$&$8.3$&$4.6$&$6.1$\\
& {pre-emp} &$6.5$&$8.7$&$4.5$&$5.7$\\
\bottomrule\end{tabular}
   \caption{Comparison of models trained with or without a learnable pre-emphasis layer. All models are initialized either with the scattering or gammatone initialization, and the pooling function is a fixed squared Hanning window.}
    \label{tab:results_preemp}
    \vskip-0.3cm
\end{table}
\endgroup

\subsection{Baseline results}
Table~\ref{tab:results} contains our results together with end-to-end baselines from the literature. \cite{amodei2016deep} is the current state-of-the-art on the WSJ dataset; it is given as a topline but uses much more training data ($\sim 12,000h$ of speech) so the results are not comparable.  \cite{graves2014towards,miao2015eesen,chorowski2016seqseq,kim2017joint} are representative results in terms of WER and LER from the literature of end-to-end models trained on speech features from 2014-2017, in chronological order. \cite{tjandra2017sequence} and \cite{tjandra2017attention} are the current state-of-the-art in LER on speech features and from the waveform respectively. These comparisons validate our baseline model trained on mel-filterbanks as a strong baseline in light of recent results, as it outperforms the state-of-the-art in LER by a significant margin ($4.9\%$ vs $6.1\%$ for \cite{tjandra2017sequence}), and achieves a test WER of $6.6\%$, better than all other end-to-end baselines (\cite{zhou2018policy} and \cite{ghahremani2016acoustic} report WER that are below our $6.6\%$ but are on easier closed vocabulary tasks).

\subsection{Instance normalization}

As described in Section~\ref{sec:instancenorm}, we evaluate the integration of instance normalization after the log-compression in the trainable filterbanks, which was not used in previous work \cite{hoshen2015speech,sainath2015learning,ghahremani2016acoustic,neilfbanks} but is used in our baseline. Figure \ref{fig:instancenorm} shows training LER as a function of the number of epochs for scattering-based and gammatone-based filterbanks models, with and without instance normalization. We can see that this normalization drastically improves the training stability of the gammatone-based model, while it moderately improves the scattering-based model. We observed a positive impact of instance normalization in all settings, and so only report as a reference the results of our implementation of a vanilla gammatone-based trainable filterbanks following \cite{hoshen2015speech,sainath2015learning}. Comparing gammatone (learnt)/gamm/max-pool without instance norm (under SOTA -- waveform) to the results of gammatone (learnt)/gamm/max-pool in Table~\ref{tab:results}, we see a significant improvements of both LER and WER due to instance normalization, with an absolute reduction in LER and WER of $1.5\%$ and $2.8\%$ respectively.

\subsection{Impact of the low-pass filter}

For low-pass filtering, we first compare the Han-fixed setting to max-pooling for gammatone-based filterbanks (as max-pooling was previously used in \cite{hoshen2015speech,sainath2015learning}), and to Han-learnt for scattering, all with instance normalization. The tendency is that the Han-fixed setting consistently improves the results in LER and WER of both trainable filterbanks. More importantly, using either an Han-fixed or Han-learnt filter when learning scattering-based filterbanks from a random initialization removes the gap in performance with the Gabor wavelet initialization that was observed in \cite{neilfbanks} where the lowpass filter was also initialized randomly.
This is an important result since carefully initializing the convolutional filters is both technically non-trivial, and also relies on the prior knowledge of mel-filterbanks. We believe the ability to use random initialization is an important first step for more extensive tuning of trainable filterbanks (e.g., trying different numbers of filters, decimation or convolution width).

Compared to the literature, replacing the max-pooling by a low-pass filter and adding an instance normalization layer leads to a $23\%$ relative improvement in LER and a $33\%$ relative improvement in WER on nov92-eval on the gammatone-based trainable filterbanks, a significant improvement compared to the existing approach \cite{hoshen2015speech,sainath2015learning}. Our models trained on the waveform also exhibit a gain in performance in LER of $22-31\%$ relative compared to the state-of-the-art end-to-end model trained on the waveform with its first 6 layers being pre-trained for mel-filterbanks reconstruction \cite{tjandra2017attention}, and outperform various end-to-end models trained on speech features, both in LER \cite{kim2017joint, tjandra2017sequence} and WER \cite{graves2014towards, miao2015eesen, chorowski2016seqseq}.

\subsection{Trainable filterbanks vs mel-filterbanks}

Comparing both trainable filterbanks with instance normalization to the log mel-filterbanks baseline, we observe that the performances of the Han-fixed settings and of the mel-filterbanks are comparable in terms of LER. However, we observe a consistent improvement in terms of WER of all trainable filterbanks. To the best of our knowledge, this is the first time a significant improvement in terms of WER relatively to comparable mel-filterbanks has been shown on a large vocabulary task under clean recording conditions. Some improvements on the clean test of the Switchboard dataset have previously been observed by \cite{ghahremani2016acoustic}, but their comparison point is MFCC rather than mel-filterbanks and the number of filters of the trainable architecture differs from their MFCC baseline.

\subsection{Adding a learnable pre-emphasis layer}

The first step in the computation of mel-filterbanks is typically the application of a pre-emphasis layer to the raw signal. Pre-emphasis is a convolution with a first-order high-pass filter of the form 
    $y[n] = x[n] - \alpha x[n-1],$
%
with $\alpha$ typically equal to $0.97$. This operation can be performed by a convolutional layer of kernel size $2$ and stride $1$, that can be plugged below time-domain filterbanks, initialized with weights $[-0.97 \quad 1]$, then learned with the network. In Table \ref{tab:results_preemp}, we compare the performance of identical models (all using a fixed Hanning window, and a gammatone or scattering initialization) with and without pre-emphasis. We observe a gain on both LER and WER (except on nov93-dev WER/scatt) when using pre-emphasis. 

\section{Conclusion}
This paper presents a systematic study of two approaches for trainable filterbanks, which clarifies good practices and identifies better architectures to learn from raw speech. Our results show that adding an instance normalization layer on top of the trainable filterbanks is critical for learning gammatone-based architectures, and speeds up learning of scattering-based architectures. Second, the use of a fixed squared Hanning window as low-pass filter is critical to learn the scattering-based filterbanks from random initialization of the filters, and improves on max-pooling for gammatone-based filterbanks. With these two improvements, we observe a consistent reduction of WER against comparable mel-filterbanks on the open vocabulary task of the WSJ dataset, in the setting of speech recognition under clean recording condition -- most likely the setting on which mel-filterbanks have been the most heavily tuned.

\section{Acknowledgements}
This research was partially funded by the European Research Council (ERC-2011-AdG-295810 BOOTPHON), the Agence Nationale pour la Recherche (ANR-10-LABX-0087 IEC, ANR-10-IDEX-0001-02 PSL*).

\bibliographystyle{IEEEtran}

\bibliography{mybib}

\end{document}